# Stem-leaf segmentation and phenotypic trait extraction of maize shoots from three-dimensional point cloud

**Chao Zhu [1], Teng Miao[1*], Tongyu Xu [1] , Tao Yang [2] , Na Li [1]**

[1] College of Information and Electrical Engineering, Shenyang Agricultural University, Shenyang, China

[2] School of Information and Intelligence Engineering, University of Sanya, Sanya, China

**\* Correspondence:**
Teng Miao
miaoteng@syau.edu.cn

**Abstract**

Nowadays, there are many approaches to acquire three-dimensional (3D) point clouds of maize plants. However, automatic stem-leaf segmentation of maize shoots from three-dimensional (3D) point clouds remains challenging, especially for new emerging leaves that are very close and wrapped together during the seedling stage. To address this issue, we propose an automatic segmentation method consisting of three main steps: skeleton extraction, coarse segmentation based on the skeleton, fine segmentation based on stem-leaf classification. The segmentation method was tested on 30 maize seedlings and compared with manually obtained ground truth. The mean precision, mean recall, mean micro F1 score and mean over accuracy of our segmentation algorithm were 0.964, 0.966, 0.963 and 0.969. Using the segmentation results, two applications were also developed in this paper, including phenotypic trait extraction and skeleton optimization. Six phenotypic parameters can be accurately and automatically measured, including plant height, crown diameter, stem height and diameter, leaf width and length. Furthermore, the values of $R^2$ for the six phenotypic traits were all above 0.94. The results indicated that the proposed algorithm could automatically and precisely segment not only the fully expanded leaves, but also the new leaves wrapped together and close together. The proposed approach may play an important role in further maize research and applications, such as genotype-to-phenotype study, geometric reconstruction and dynamic growth animation. We released the source code and test data at the web site https://github.com/syau-miao/seg4maize.git

## 1    Introduction

Population growth and climate change are expected to be the most pressing challenges to food availability for most of the current century and further in the future (Lenaerts et al., 2019; Hickey et al., 2019). Plant genotyping and phenotyping technologies are crucial for the acceleration of breeding programmes to meet increasing food and energy demands (Tester et al., 2010; Zhao et al., 2019). High-throughput plant phenotyping has been identified as a bottleneck for crop improvement (Araus et al., 2018). The conventional method for the evaluation of phenotypic traits is tedious, destructive, time-consuming, labour-intensive, and error-prone, which affects the dynamic monitoring of plant development throughout the growth period (Pajares et al., 2011; Chen et al., 2011; Pan et al., 2007) Therefore, it is essential to develop an automatic, accurate, and non-invasive method for phenotypic trait extraction in plants (Yang et al., 2013), especially at organ level (Speck et al., 2011; Bertin et al., 2009).

Due to the widespread availability of digital cameras, two-dimensional (2D) images have been extensively used in image-based phenotyping techniques for analysing canopy cover (Brichet et al., 2017), phenotypic parameters of barley (Paulus et al., 2014) and more. However, 2D imaging cannot be used for measuring several traits such as total area, mean leaf angle and volume, due to the complex three-dimensional (3D) structure of plants (An et al., 2017).

A feasible technique for extracting morphological traits of plants is based on 3D point clouds, which are obtained by digitization of plants. Several methods have been proposed to obtain complete 3D point clouds of crops, including the utilization of structure from motion (Hartley et al., 2003), stereo vision (Bao et al., 2019; Wu et al., 2020), depth cameras (An et al., 2017; Azzari et al., 2013; Sun et al., 2017), time of flight cameras (Alenyà et al., 2011; Paulus et al., 2014), light detection and ranging (LiDAR) (Garrido et al., 2015; Tang et al., 2014; Eitel et al., 2014; Li et al., 2018) and more. An indication of the complexity of plant point clouds is found in recent research on 3D point-cloud-based phenotyping pipelines, in which the 3D point cloud was still segmented manually (Duan et al., 2016; Zhu et al., 2020). Several recent studies have focused on automatic organ segmentation and trait extraction of maize using point clouds（Jin.et al,2018;Wu et al.,2019; Elnashef et al.,2019）. Wu et al. (2019) used the Laplacian contraction method to extract the 3D skeletons of maize plants and phenotypic traits such as leaf inclination and stem length. Their method is very robust for fully developed leaves. However, it is difficult to extract the skeletons of new emerging leaves that are very close to each other. This is because the Laplacian contraction method shrinks the point cloud at the base of these new leaves, in error, into the same sub-skeleton which was classified as stem. Elnashef et al. (2019) proposed a 3D point cloud classification and segmentation algorithm based on tensors and realized organ-level classification of various crop species by using first- and second-order tensor analysis. To segment individual leaves this method used the density-based spatial clustering of applications with noise (DBSCAN) algorithm (Ester et al., 1996), which can reliably separate distinct organs from each other. However, for the maize plant, the leaf base of the larger new leaf surrounds the smaller leaf; hence, it is difficult to find a suitable set of DBSCAN parameters to separate these leaves correctly. Jin et al. (2018) developed a median normalized-vector growth algorithm for stem-leaf segmentation. This method marks the bottom of the stem in the point cloud manually, then uses the growth region to segment first the stem and then the individual leaf, in order from leaf base to tip. The method is very simple and efficient, and the segmentation effect is excellent for the middle and lower leaves, but for the new leaves, false segmentation will occur.

The existing stem-leaf methods based on point clouds are more effective for maize plants with fully expanded leaves. However, it is difficult to separate the new emerging leaves using these methods because these leaves are often close together and wrap together. Further, some methods require human interaction to achieve better stem-leaf segmentation. Therefore, the automatic and fine segmentation of seedling maize is still a critical challenge.

To address this issue, we propose a segmentation algorithm, based on 3D point clouds, for stems and leaves of maize plants to achieve automatic, high-precision, and high-throughput extraction of the morphology of these plants. The main contributions of our study are as follows: (1) we provide a fully automatic method for stem-leaf segmentation based on skeleton information; (2) the proposed method is very effective for the situation in which multiple new emerging leaves are close to each other and larger new leaves wrap around smaller leaves.

## 2    Materials and methods

### 2.1    Experimental material

The field experiment was conducted from May to July 2019 at the experimental maize field of Shenyang Agricultural University (39°56′N, 116°16′E). The tested maize variety was XianYu 335, which was planted in a plot with an area of 666 m². The maize row distance was 60 cm and the plant spacing was 25 cm. To ensure the morphological stability of the leaves of these plants during the acquisition of 3D point cloud data, the plants were transplanted into pots with their underground sections and watered immediately. These pots were then transported to an indoor laboratory. Thirty individual maize samples were chosen from vegetative stages V3 to V8 for analysis.

### 2.2    Data acquisition platform and data preprocessing

A high-accuracy, high-resolution, non-destructive image system is designed for data acquisition, it was mainly composed of a computer station(Dell, Inc., Texas, American), a hand-hold laser scanner(Beijing Tenyoun 3d Technology Co., Ltd. ,Beijing, China) and a movable stand with laser scanning points(Figure 1). The hand-hold laser scanner's data acquisition is based on laser triangulation principle, its weight is 0.95kg, the accuracy is as high as 0.030mm, and the scanning rate is 240000 times / s, a more detailed description of the hand-hold laser scanner system specifications is shown in Table 1.

Combined with the FreeScan software platform of 3D scanner, the point cloud data of maize plant can be obtained quickly and accurately by multi view scanning around the plant. As a result, 50000 to 100000 points were obtained for every maize plant. Due to the influence of external environment and equipment itself, some noise points are produced in the process of obtaining point cloud data. To improve the accuracy of the model and the data processing speed, we used the CloudCompare software to denoise and down sample the raw point cloud data. After down sampling, the number of point clouds per plant used in this paper was about 5000 to 8000. We experimentally confirmed that point clouds of this scale retained the morphological characteristics of maize plants, and was suitable to obtain better skeleton efficiently.

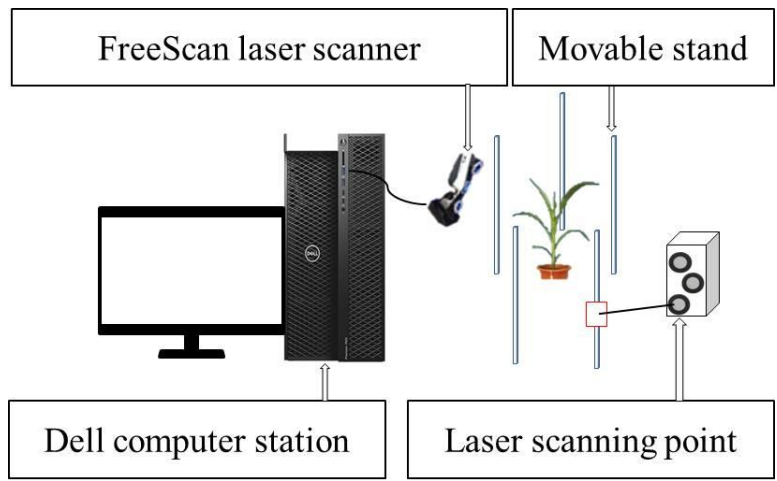

**Figure 1.** The composition of 3D laser scanning system

**Table 1.** General specification of FreeScan laser scanner

| Specification | Value |
|---|---|
| Weight | 0.95kg |
| Speed point/s | 240000 |
| Laser category | II |
| Scanning range | 280×250mm |
| Scanning accuracy | 0.030mm |
| Working distance | 300mm |
| Depth of field | 250mm |
| Range | 0.1-6m(scalable) |
| Resolution | 0.100mm |
| Software | FreeScan |
| Operating temperature | -10 to +45℃ |
| Interface standard | USB 3.0 |

## 2.3 Steps of the proposed method

Our stem-leaf segmentation method involves three steps: (1) skeleton extraction; (2) coarse segmentation based on the skeleton; (3) fine segmentation based on stem-leaf classification. Using the segmentation results, two applications were developed, including phenotypic trait extraction and skeleton optimization. The overall workflow is shown in Figure 2.

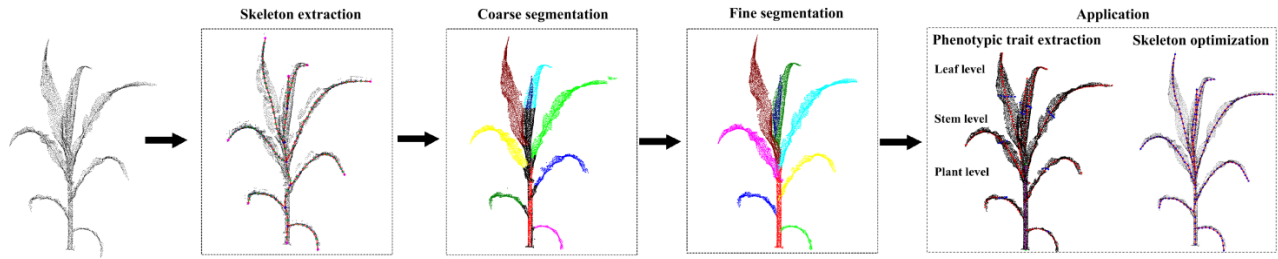

**Figure 2.** Workflow of the proposed automatic stem-leaf segmentation method.

### 2.3.1 Skeleton extraction

In this paper, the Laplacian based method (Cao et al., 2010) was used to generate a plant skeleton for given maize point cloud $p = \{p_i\}$. The data structure of the plant skeleton consisted of three parts: (1) a skeleton vertex set $U = \{u_i\}$; (2) an undirected graph $G$ whose vertices represented the skeleton vertex set $U$; (3) for each skeleton vertex, a corresponding point set which was a subset of the initial point cloud $p$. Suppose the plant skeleton has $N$ vertices, and the corresponding point set for the i-th skeleton vertex $u_i$ is $C_i$, then $C_1 \cup C_2 \cup ... C_N = p$, $C_i \cap C_2 = \emptyset$. Each skeleton vertex $u_i$ was generated from the corresponding point set $C_i$ using the Laplacian method. We need to emphasize the parameter values used in the Laplacian based method. We found by experiment that the default parameters used by Cao et al. (2010) can provide a good skeleton for most plants. However, the parameter k used to construct the Laplacian matrix, can affect the skeleton extraction of new emerging leaves. k is defined as the number of nearest neighbors required to construct the domain over which the tangential plane is estimated. As the value of k increases, more local features of the point cloud are ignored in the skeleton. The choice of k value will be discussed in Section 3.1.1.

### 2.3.2 Coarse segmentation based on skeleton

A skeleton vertex with one neighbor, two neighbors, or more than two neighbors is called a root vertex, branch vertex, or junction vertex, respectively, as shown as in Figure 3(A). It is clear that only one root vertex is generated by the stem point cloud (called the stem root vertex), and the remaining root vertices are generated by leaf-tip point clouds (called tip root vertices). The key problem of skeleton segmentation is to recognize the only stem root vertex from all the root vertices.

We used a morphological feature of maize leaf to identify the stem root vertex. We found that the new leaves of maize are upward when growing. When some mature leaves are fully expanded, the upper part may be downward, but the lower part may still be upward. Based on this observation, we used the following methods to select the stem root vertex (Figure 3(A)).

Step 1: We calculated all the shortest paths between a root vertex and its nearest junction vertex. In this paper, this path is called R-J sub-skeleton.

Step2: Sorted the vertices of each R-J sub-skeleton. When sorting, we took root vertex as the first vertex, and arranged the rest vertices in order according to the connection of edges.

Step 3: After sorting, for each R-J sub-skeleton, the vertex $v_1$ at the middle position and the vertex $v_2$ at the last position were selected. We calculated a direction vector $\vec{v} = \frac{v_1 - v_2}{\|v_1 - v_2\|}$ for each R-J sub-skeleton.

Step 4: If the direction vector of the i-th R-J sub-skeleton was $\vec{v}^i$ and the direction vector of the j-th R-J sub-skeleton was $\vec{v}^j$, then the angle between the two sub-skeletons was defined as $\vec{v}^i \cdot \vec{v}^j$. For any R-J sub-skeleton, we calculated its angle with other R-J sub-skeletons and found the sum of all the angles. The root vertex of the R-J sub-skeleton with the smallest sum was the stem root vertex, and the rest were the tip root vertices.

 The reason why this algorithm works well is that if the R-J sub-skeleton is a leaf skeleton, its direction vector indicates the direction of the lower part of the blade, which is upward. If the R-J sub-skeleton is a stem skeleton, its direction vector is downward after sorting. Therefore, the angle between the stem sub-skeleton and other leaf sub-skeletons will appear a large number of negative values, and the angle sum is the minimum.

In order to determine the approximate position of the point in the plant through three-dimensional coordinates, we transformed the plant coordinate from the initial global coordinate system to a plant coordinate system. This made the subsequent segmentation algorithm more concise. First, we estimated the growth direction of the maize plant. The growth direction is defined as a vector that coincides with the median axis of the stem, the positive direction being from lower to upper part of the plant. To obtain the growth direction, we first found the shortest path between the stem root vertex and its farthest junction vertex in the skeleton graph G. Then, we used the skeleton points in the shortest path to fit a straight line, which approximately represented the growth direction. After obtaining the growth direction, we constructed a plant coordinate system, with the growth direction as the Z axis. To determine the other two axes, we projected the point cloud onto a plane where the growth direction was the normal vector. Next, we used principal component analysis (PCA) to determine the first and second principal component vectors, and defined these as the X and Y axes, respectively (Figure 3(B)). All subsequent algorithms were simplified by being performed in the plant coordinate system.

After finding the stem root vertex, the plant skeleton was decomposed into one stem sub-skeleton and several leaf sub-skeletons (Figure 3(C)). The leaf sub-skeleton was the shortest path from a tip root vertex to its nearest junction vertex in the plant skeleton graph G. The stem sub-skeleton denoted the shortest path from the stem root vertex to its farthest junction vertex in G. We used the Dijkstra method (Dijkstra,E.W.,1959) for finding the shortest paths. With these sub-skeletons, we obtained an initial (coarse) result of stem-leaf segmentation. For any sub-skeleton, we combined the corresponding point sets of its vertices together to obtain an initial(coarse) segmentation result of the organ instance represented by this sub-skeleton.

 To facilitate further stem-leaf segmentation, we established an unsegmented point cloud set $\emptyset_u$, a stem point cloud set $\emptyset_s$, and several leaf point cloud sets $\emptyset_l^i$ (i=1,2……m), where m is the number of leaves. The corresponding point cloud set of all the skeleton vertices of the i-th leaf sub-skeleton represented the segmented points for the i-th leaf instance; hence, we placed these points in the corresponding leaf point cloud set $\emptyset_l^i$ (represented by colored dots in Figure 3(C)). The coarse segmentation results of other organ instances were also obtained in this way.

Because the corresponding point cloud of the stem skeleton contained many leaf points, they were initially placed into the unsegmented point cloud set $\emptyset_u$ (represented by black dots in Figure 3(C)). We then found some points in the set $\emptyset_u$ that must belong to the stem points set $\emptyset_s$. First, we used the stem sub-skeleton to find some points which were not stem points with high probability. We observed that in the stem skeleton, the sub-skeleton between the last two junction vertices was mainly formed by the new leaf point; hence, we used the minimum Z coordinate of the point cloud of this sub-skeleton as a truncation threshold. The point clouds of all the stem skeletons with z coordinates higher than the truncation value were kept in $\emptyset_u$, and the remaining points were used to select the stem points by a radius constraint. The constraint was that the stem points were required to reside within radius $\alpha r_c$ of the central axis of the stem. The parameter $r_c$ can be calculated by the point clouds in $\emptyset_u$. $\alpha$ ($\alpha \geq 0$) is a user tuning parameter to scale the parameter $r_c$, the choice of $\alpha$ value will be discussed in Section 3.1.1. To calculate  $r_c$, we first selected those points from $\emptyset_u$ whose z coordinate was less than $\frac{z_m}{2}$, where $z_m$ was the median z coordinate of all the point clouds in $\emptyset_u$, and fitted the central axis of the stem. These points were chosen because the point cloud of the lower stem contained fewer leaf point clouds. We used the least squares method to fit the central axis of these points and calculated the median distance of all the points from this central axis as parameter $r_c$. Point clouds with distances smaller than $\alpha r_c$ were removed from set $\emptyset_u$ and added to stem point set $\emptyset_s$ (represented by red dots in Figure 3(C)).The resulting coarse segmentation, obtained after segmenting the stem points using the above three constraints, is shown in Figure 3(D).

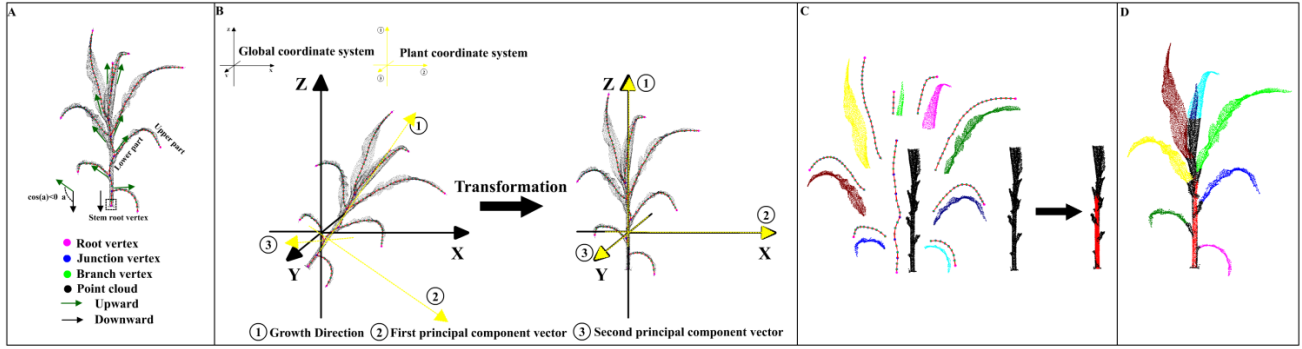

**Figure 3.** Process flow of coarse segmentation based on plant skeleton. (A) Finding the stem root vertex. (B) Transforming the cloud point from global coordinate system to plant coordinate system. (C) Decomposition of plant skeleton into sub-skeletons and selecting the stem points (D) Result of coarse segmentation.

### 2.3.3 Fine segmentation based on stem-leaf classification

We then performed fine segmentation on the above coarse segmentation result. The main objective of fine segmentation was to classify the points of $\emptyset_u$ into the leaf point sets $\emptyset_l^i$ and stem point set $\emptyset_s$ using the stem-leaf classification method (Figure 4). We took out points from $\emptyset_u$ in turn to determine which organ set this point belongs to. The order in which points were taken from $\emptyset_u$ and the way for judging which organ a point belonged to directly affected the quality of our classification results. First, we introduce how to determine the order. We found that most of the unsegmented points belong to the new leaves. Therefore, in order to ensure that the new leaf instance get their points as completely as possible, the order of selecting points from $\emptyset_u$ was to select the upper point first, and then the lower point. Next, we briefly introduce the principles of organ classification for a point. The first principle was the spatial consistency between a point and an organ. In short, the closer a point was to which organ, the greater the probability that it belonged to the organ. The second principle was the geometric consistency principle, which was specific to the leaf instance. We observed that the maize leaves are flattened in parts, so if a point belonged to a leaf instance, the point should be coplanar with the neighboring points in the instance.

The specific classification process is described as follows:

Step 1. According to the principle of the point selection order, the points in the set $\emptyset_u$ were arranged in descending order according to their z coordinates. The points from the sorted set $\emptyset_u$ were used in the following steps.

Step 2. After selecting a point p from $\emptyset_u$, the corresponding organ point cloud set was determined, represented by $\emptyset'$. The average Euclidean distance $C_1$ from point p to any point set $\emptyset_v$ was defined as follows:

$$C_1 = \frac{1}{K_1}\sum_{p(j)\in A}\|p(j) - p\|_2, \tag{1}$$

where $\| \quad \|_2$ denotes $L_2$ distance between p and its $K_1$ nearest neighbours p(j) (represented by set A) in set $\emptyset_v$ .This distance $C_1$ was calculated from point p to all the organ sets, including all the leaf point cloud sets and the stem point cloud set. The organ sets $\emptyset_v^1$ with the smallest distance $\boldsymbol{C_1^1}$ and $\emptyset_v^2$ with the second smallest distance $\boldsymbol{C_1^2}$ were determined. If $\frac{(c_1^2 - c_1^1)}{c_1^1} > \boldsymbol{\sigma}$ (in this paper, $\sigma$ is set to 0.2), then $\emptyset' = \emptyset_v^1$, and step 3 was directly executed. Otherwise, the distance C between point p and one of the two sets was calculated as follows:

$$C = C_1 + \beta C_2, \tag{2}$$

$$f(x, y, z) = n_x x + n_y y + n_z z + d, \tag{3}$$

$$C_2 = \frac{|n_x p_x + n_y p_y + n_z p_z + d|}{\sqrt{n_x n_x + n_y n_y + n_y n_y}}. \tag{4}$$

$C_2$ is the distance from point p to a local plane $f(x, y, z)$, which was fitted to the $K_1$ nearest neighbours of p in set $\emptyset_v$ using the least squares method. $\beta$ is a user adjustment parameter, which is used to adjust the proportion of $C_2$ in C. The choice of β value will be discussed in Section 3.1.1. $C_1$ denotes the Euclidean distance between point p and set $\emptyset_v$, and $C_2$ denotes the point-to-plane distances between p and $\emptyset_v$. The values of C from point p to $\emptyset_v^1$ and $\emptyset_v^2$, were calculated and the set with the smaller distance is denoted as $\emptyset'$.

Step 3. The point p was removed from $\emptyset_u$ and placed into $\emptyset'$. The updated $\emptyset'$ was used in the next point classification process (steps 2 and 3). Therefore, in the classification process, the elements in $\emptyset_u$ gradually decreased, while those in $\emptyset_s$ and $\emptyset_l^i$ gradually increased. When the set $\emptyset_u$ was empty, the fine segmentation process was completed.

The $C_1$ term represents the principle of the spatial consistency. It is obvious that the smaller the distance, the greater the probability that the point belongs to that organ. When the distance between the point and the two organs was relatively close, it was difficult to determine the correct category of the point only by using item $C_1$. In this case, we added item $C_2$ to help classification. The $C_2$ term represents the geometric consistency principle. The inclusion of the $C_2$ term helps to deal with the situation in which large leaves wrap small leaves and the small leaves are close to the large leaves. Figure 5(A) shows segmentation results without $C_2$ term. Here, the point cloud in the black circle, which should belong to the large leaves, is segmented into small leaves. Figure 5(B) shows the segmentation results after adding the $C_2$ term. The advantage of including the $C_2$ term is evident in this figure.

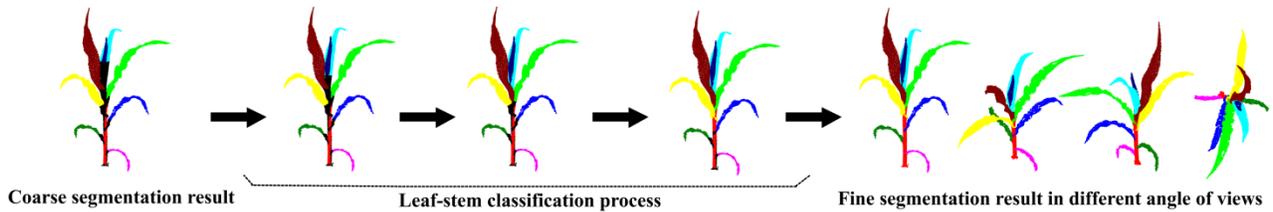

Coarse segmentation result        Leaf-stem classification process        Fine segmentation result in different angle of views

**Figure 4.** Process flow of fine segmentation based on stem-leaf classification.

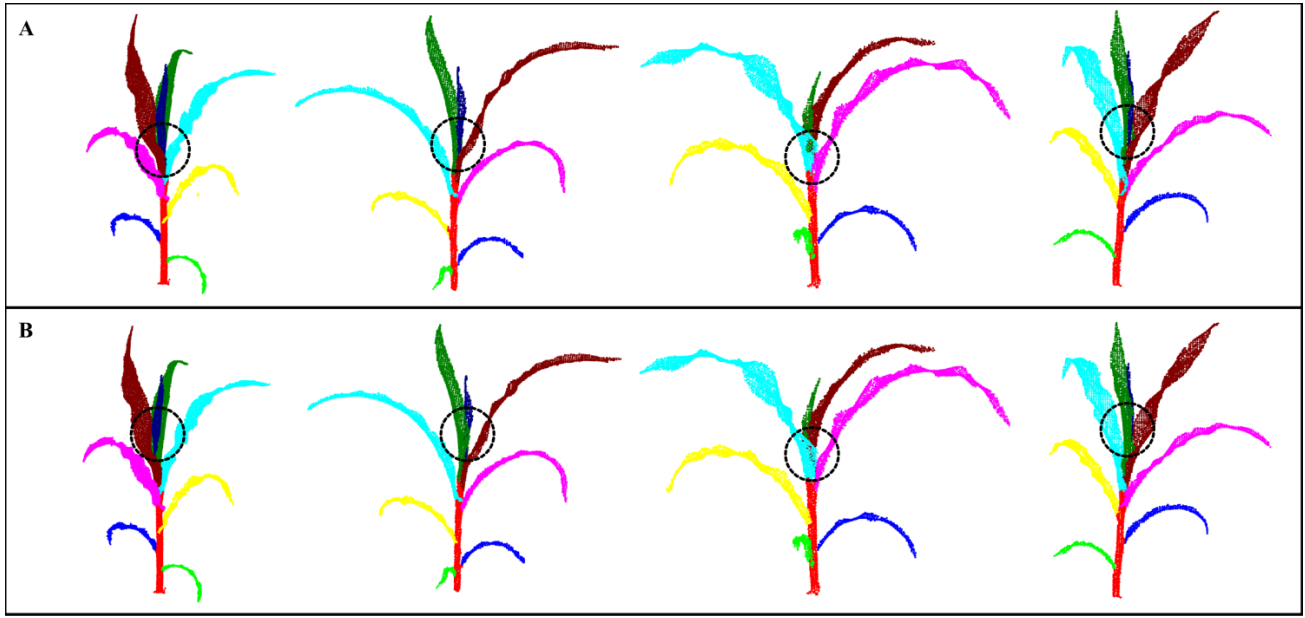

**Figure 5.** Effect of $C_2$ term on the segmentation results. Segmentation results (A) without $C_2$ term and (B) with $C_2$ term.

## 2.4 Application

### 2.4.1 Phenotypic trait extraction

The segmented leaf and stem instances were used to extract phenotypic traits at the levels of individual maize plant, stem, and leaf (Figure 6). The difference between the maximum and minimum z values of the point cloud yielded the plant height, and the difference between the maximum and minimum x values yielded the crown diameter. Using the two traits, an indirect trait CHR was calculated through dividing crown diameter by plant height, which can depict the compactness of the maize.

At stem level, we fitted a straight-line segment to the stem points using the least squares method. The length of this line segment was equal to the stem height. We calculated all the projection distances from the stem points to this line segment and used twice the median of these distances as the stem diameter.

At the leaf level, we first conducted PCA to compute first, second and third principal component axes. We then determined the two end points along the first axis. The shortest path between these two end points was obtained. This was the "length path", whose length was equal to the leaf length. To calculate the leaf width, we first decomposed the leaf point cloud into n (here, n = 11) intervals along the first principal component vector. Then we found two end points along the second principal component axis and two end points along the third principal component axis in the point cloud of the middle interval. We determined two shortest paths between two groups of end points, the "width path" was the longest path, whose length was equal to the leaf width. Figure 6 shows the length paths and width paths of three different leaves.

Automatic stem-leaf segmentation and phenotypic trait extraction of maize shoots at seedling stage from three-dimensional point cloud

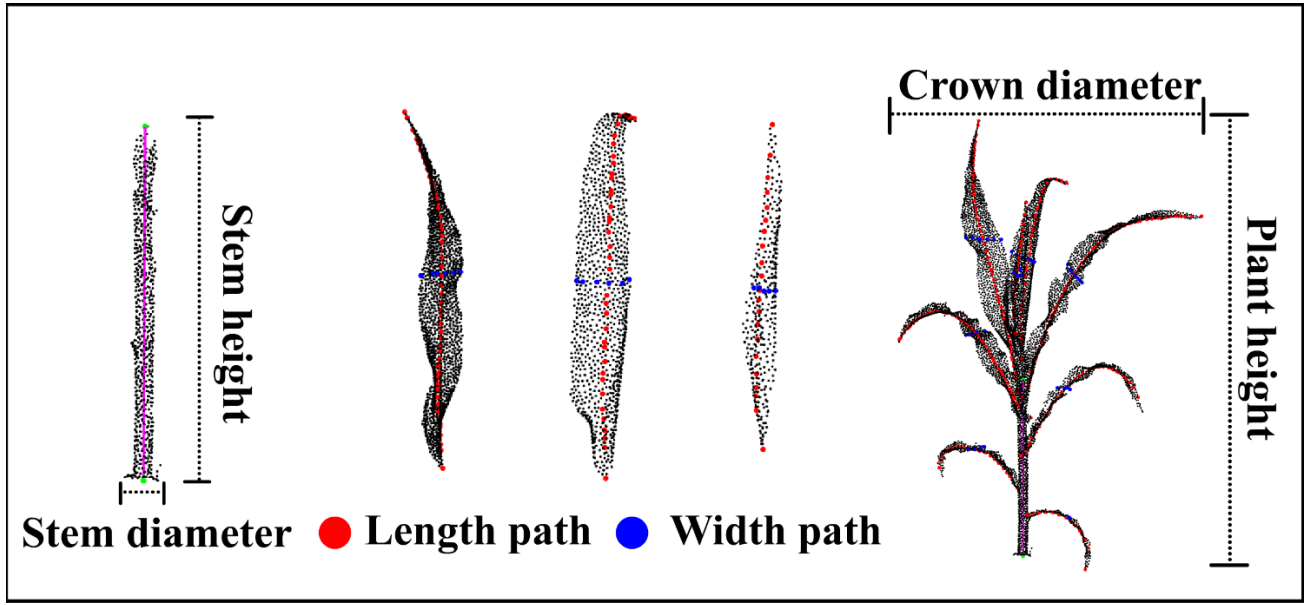

**Figure 6.** Structure and phenotypic traits of the individual maize in leaf, stem, and plant levels.

### 2.4.2 Skeleton optimization

We can segment the maize point cloud based on the skeleton generated by Laplacian based method, and conversely, we can optimize the initial skeleton according to the segmentation result. A key problem of the initial plant skeleton generated by Laplacian based method was that the point clouds of different organs may form a skeleton vertex together, so it was difficult for the plant skeleton to be decomposed into a complete organ skeleton conforming to the morphology of each organ. The reason for this problem is that in the skeleton extraction algorithm, the farthest-point sampling method was used to calculate skeleton vertices. It roughly classified all points within a certain distance into a group(As shown in figure 7A), and calculated the barycenter of these points as a skeleton vertex, regardless of which organ they actually belonged to. We proposed an optimization method to solve this problem as follows:

Step 1. We judged which organ each skeleton vertex belongs to according to the segmentation result of the point cloud. If all the corresponding point clouds of a skeleton vertex belonged to the same organ *i*, then this skeleton vertex was considered as a skeleton vertex of organ *i*. If its corresponding point clouds contained points of multiple organs, then we considered this vertex to be a composite skeleton vertex. (As shown in figure 7B)

Step 2. We decomposed each composited skeleton vertex into several vertices. If the corresponding point clouds of a composited skeleton vertex contained the point cloud of N organs, then they were segmented into N groups according to their categories, and the barycenter of each point group was the new skeleton vertex of the corresponding organ. (As shown in figure 7C)

Step 3. The skeleton vertices of each organ were connected to form an organ skeleton (As shown in figure 7D). The most terminal skeleton point in the leaf skeleton was connected with the nearest skeleton vertex on the stem.

Step 4. As the maize stem is relatively straight, the stem skeleton should be calibrated. Stem skeleton vertices were used to fit a straight line segment based on least square distance, and then they were projected on the new fitted line segment.

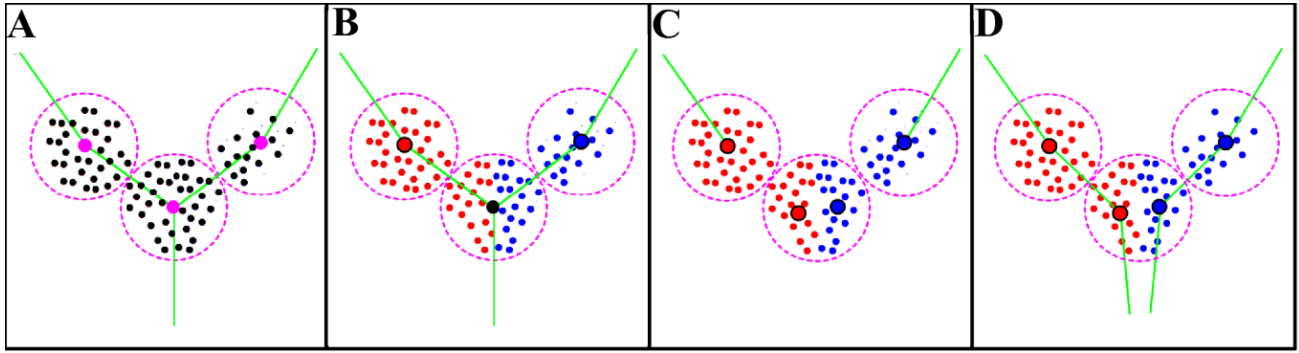

**Figure 7.** Skeleton optimization diagram. (A) represents the Laplacian based method roughly classified all points within a certain distance into a group, and calculated the barycenter of these points as a skeleton vertex. Black dots represent plant point clouds. The pink dots represent the skeleton vertex. The black dots inside a pink dotted circle are the corresponding point cloud of the skeleton point (pink dot) .The green line is the edge. (B) represents the point clouds in a group may belong to different organs. The small red dots and the small blue dots belong to two different organ instances. Black dot represents a composite skeleton vertex, and its corresponding point clouds belong to two organ instances (red and blue). (C) represents the decomposition of the composite skeleton vertex into two skeleton vertices of different instances. (D) represents the new skeleton vertices were connected into the plant skeleton.

## 2.5 Statistical Analyses

To evaluate the accuracy of the stem-leaf segmentation algorithm, 30 selected plants of different vegetative stages of maize seedling, with different numbers of leaves , different height, different CHR for quantitative analysis. The segmentation results were compared with the ground truth obtained manually using the software MeshLab (Cignoni et al., 2008), as suggested by Elnashef et al(2018) . We calculated the precision, recall, f1 score for each stem and leaf instance. The precision, recall, micro f1 score and overall accuracy were also calculated for the individual maize.

Linear regressions were used for comparisons of the calculated phenotypic traits and the ground truth measured manually. The estimated coefficient of determination ($R^2$), root-mean-squared error (RMSE) were calculated to assess the accuracy of these extracted traits.

## 3 Results

The algorithm described in Section 2 was implemented in MATLAB on a computer with Core i7 processor and 32 GB memory. The average time of 30 tested data required for skeleton extraction, and segmentation was approximately 11.48 s. About 85% of the time was spent on skeleton extraction.

## 2.4 Results for Stem-leaf segmentation

The representative segmentation results were shown in Figure 8, i.e. , the results of the highest overall accuracy, lowest overall accuracy , highest micro f1 score, lowest micro f1 score, largest leaf number, smallest leaf number , highest height, lowest height, largest CHR and smallest CHR. In each subplot, the left and right figure were the algorithm results and ground truth. To validate our algorithm's ability to segment new leaves, the segmentation results from different view angles are shown in Figure 9. Figure 10 showed overall accuracy and micro f1 score of individual plant under different leaf numbers, heights and CHR values. The results showed that our method generated similar results as the ground truth no matter the height was low or high, the structure was compact or flat, the leaf number was less or more. Coincidentally, the highest overall accuracy and highest micro F1 score appeared at the same sample, the lowest overall accuracy and lowest micro F1 score appeared at the same sample. The mean precision, mean recall, mean micro F1 score and mean over accuracy of each individual maize were 0.964, 0.966, 0.963 and 0.969 (table 2)

In order to evaluate the segmentation results of different organ instances, we proposed an instance location variable $L$ to represent the relative position of organs on the plant. Take a maize plant with $N$ leaves as an example. The stem instance was represented by $L=0$, and the i-th leaf instance from bottom to top was represented by $L=i/N$. The larger the $L$ value of a leaf instance, the later the leaf was born. The $L$ value of the latest newly born leaf on each plant was 1.0. Figure 10 shows the precision, recall and f1 score values of the organ instance with different instance location variable $L$. It can be seen from the figure that the segmentation results of the leaves have little difference in different leaf positions, which showed that our method was effective for fully expanded leaves and undeveloped leaves. The segmentation effect of this method for stem organs is worse than that of leaves. The mean values of precision, recall, micro F1 score of stem instance and leaf instance were 0.871, 0.868, 0.861, 0.975, 0.979, 0.976 respectively (table 2).

.In the supplementary, we showed the segmentation results of all 30 samples in detail, including the visualization results, accuracy, recall, micro f1 score, overall accuracy and the algorithm parameters used of each sample.

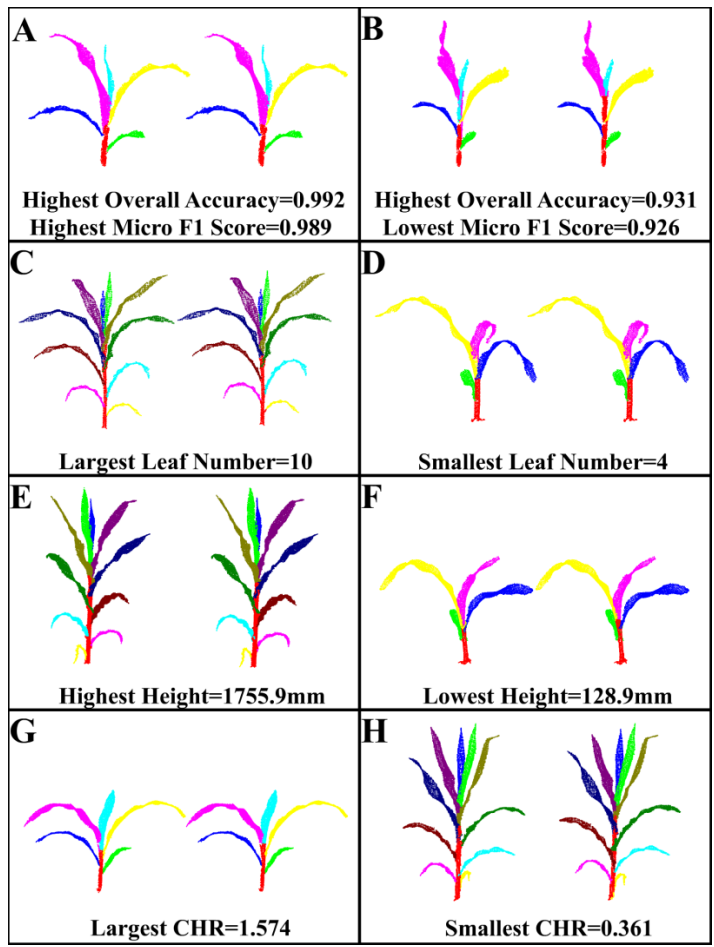

**A** Highest Overall Accuracy=0.992
Highest Micro F1 Score=0.989

**B** Highest Overall Accuracy=0.931
Lowest Micro F1 Score=0.926

**C** Largest Leaf Number=10

**D** Smallest Leaf Number=4

**E** Highest Height=1755.9mm

**F** Lowest Height=128.9mm

**G** Largest CHR=1.574

**H** Smallest CHR=0.361

Figure 8. Segmentation results of the selected maize samples. (Left) Result segmented by our algorithm. (Right) Ground truth segmented manually. Different colors represent different segmented instances, and the same leaf or stem in both ground truth and the automatic segmented results are showed in the same color. Results (A)–(H) were selected based on different properties annotated in the subfigures.

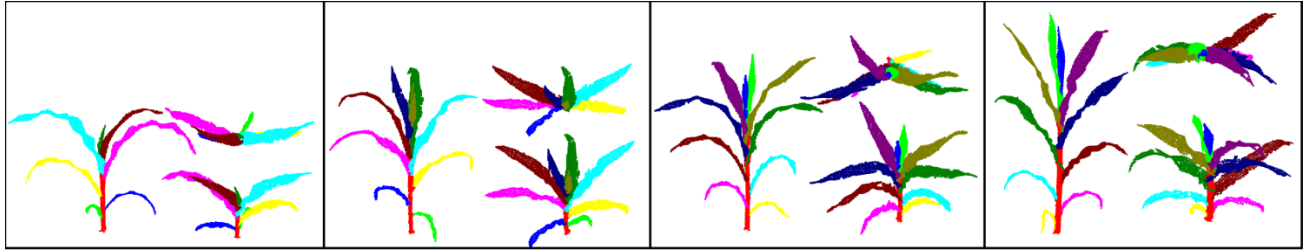

**Figure 9.** Segmentation results from different view angles.

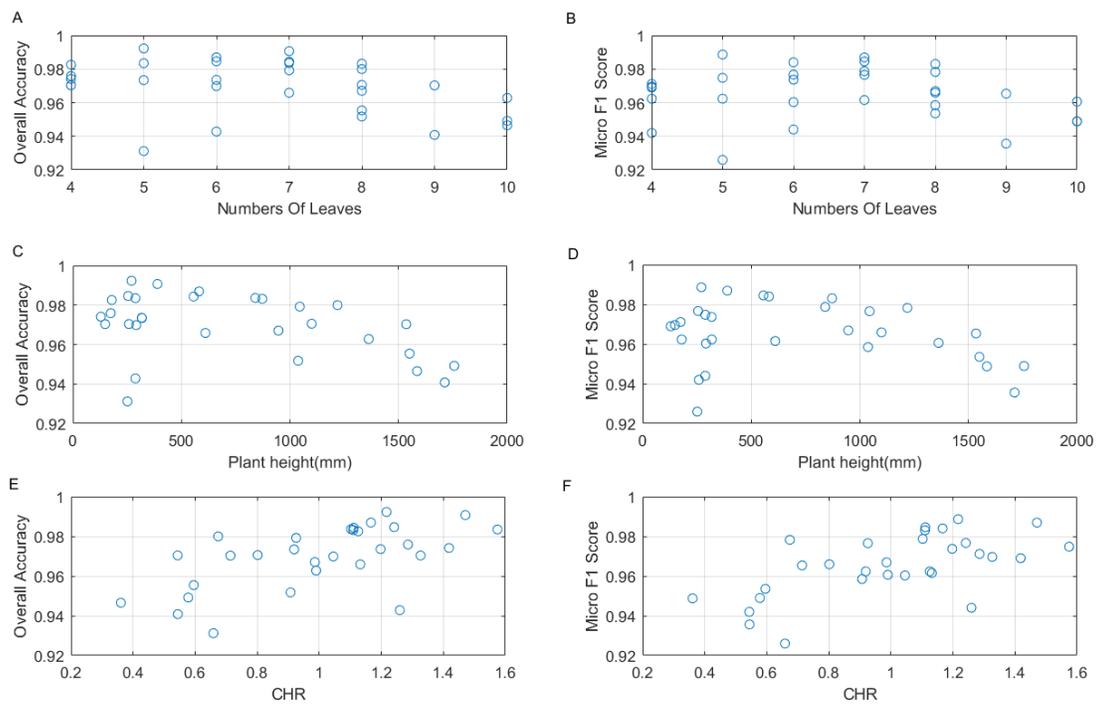

Figure 10. Accuracy assessments of the segmentation results of the 30 maize samples. (A)，(C)，(H) represent the relationship between overall accuracy and number of leaves, plant height and CHR，respectively. (B), (D), (F) represent the relationship between micro f1 score and number of leaves, plant height and CHR, respectively.

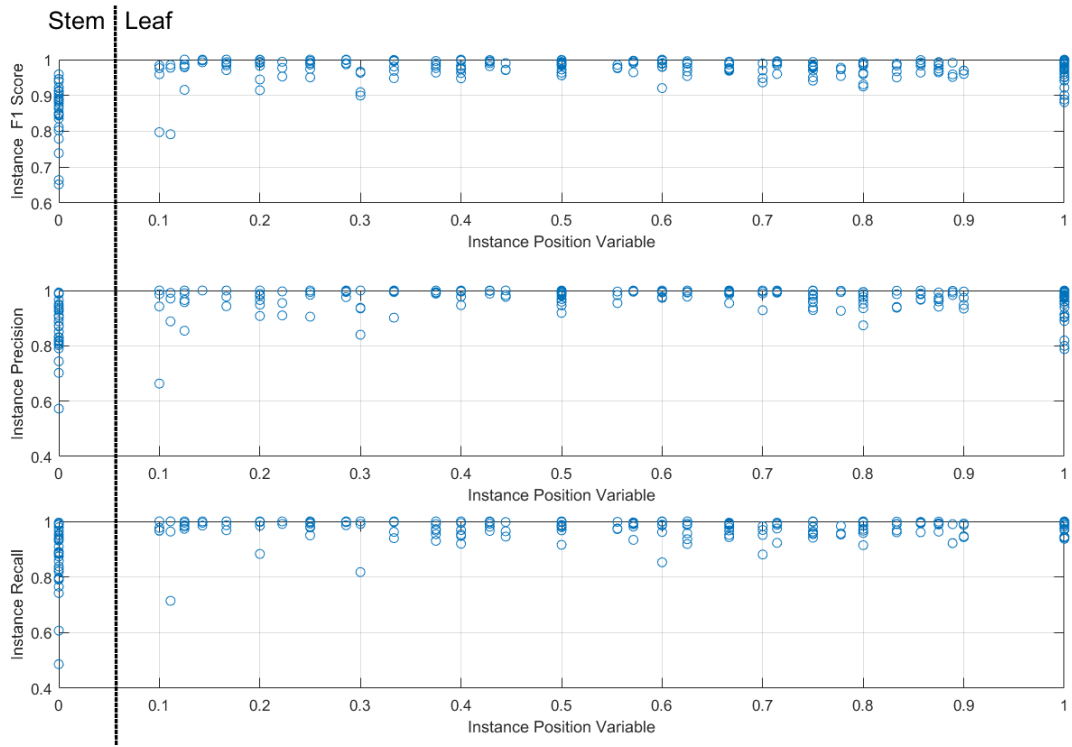

Figure 11. The accuracy of the segmentation results of all the organ instances with different instance position variable.

**Table 2.** Accuracy assessments of the segmentation results of the stem instances, leaf instances, all the organ instances and the Individual maize plants.

|  | F1 Score / Micro F1 Score | | | Precision | | | Recall | | | Overall Accuracy | | |
|---|---|---|---|---|---|---|---|---|---|---|---|---|
|  | Min | Mean | Max | Min | Mean | Max | Min | Mean | Max | Min | Mean | Max |
| Stem Instance | 0.652 | 0.861 | 0.959 | 0.573 | 0.871 | 0.993 | 0.466 | 0.868 | 0.996 | ---- | ---- | ---- |
| Leaf Instance | 0.792 | 0.976 | 1.000 | 0.664 | 0.975 | 1.00 | 0.714 | 0.979 | 1.000 | ---- | ---- | ---- |
| Organ Instance | 0.652 | 0.962 | 1.000 | 0.573 | 0.962 | 1.00 | 0.466 | 0.965 | 1.000 | ---- | ---- | ---- |
| Individual | 0.926 | 0.963 | 0.989 | 0.935 | 0.964 | 0.988 | 0.911 | 0.966 | 0.992 | 0.931 | 0.969 | 0.992 |

## 2.5    Results for Phenotypic trait extraction

Comparisons of system-derived measurements and manual measurements of six phenotypic traits (plant height, crown diameter, stem height, stem diameter, leaf width, leaf length) are shown in Figure 12.

At plant level, we evaluated 30 individual maize samples with different vegetative stages, the correlation coefficient of our method with the ground truth results is 0.99 for plant height and 0.99 for crown diameter with RMSE of 1.71 cm and 3.44 cm, respectively.

At stem level, there are total 30 segmented stems were calculated, the measured traits include stem height and stem diameter. The estimated values of these traits based on the proposed method are strongly correlated with the ground truth measurements with a correlation coefficient of 0.96 and 0.96, respectively. The RMSE of stem length and stem diameter is 7.05 cm and 0.45cm, respectively.

At leaf level, 237 segmented leaves of the 30 maize plants were extracted by our proposed segmentation algorithm, the measured leaf traits include leaf width and leaf length. For all the measurable leaves, a high correlation was observed between our results and the ground truth. The estimated $R^2$ and RMSE of leaf width (leaf length) are 0.94 and 0.77cm (0.97 and 4.75cm), respectively.

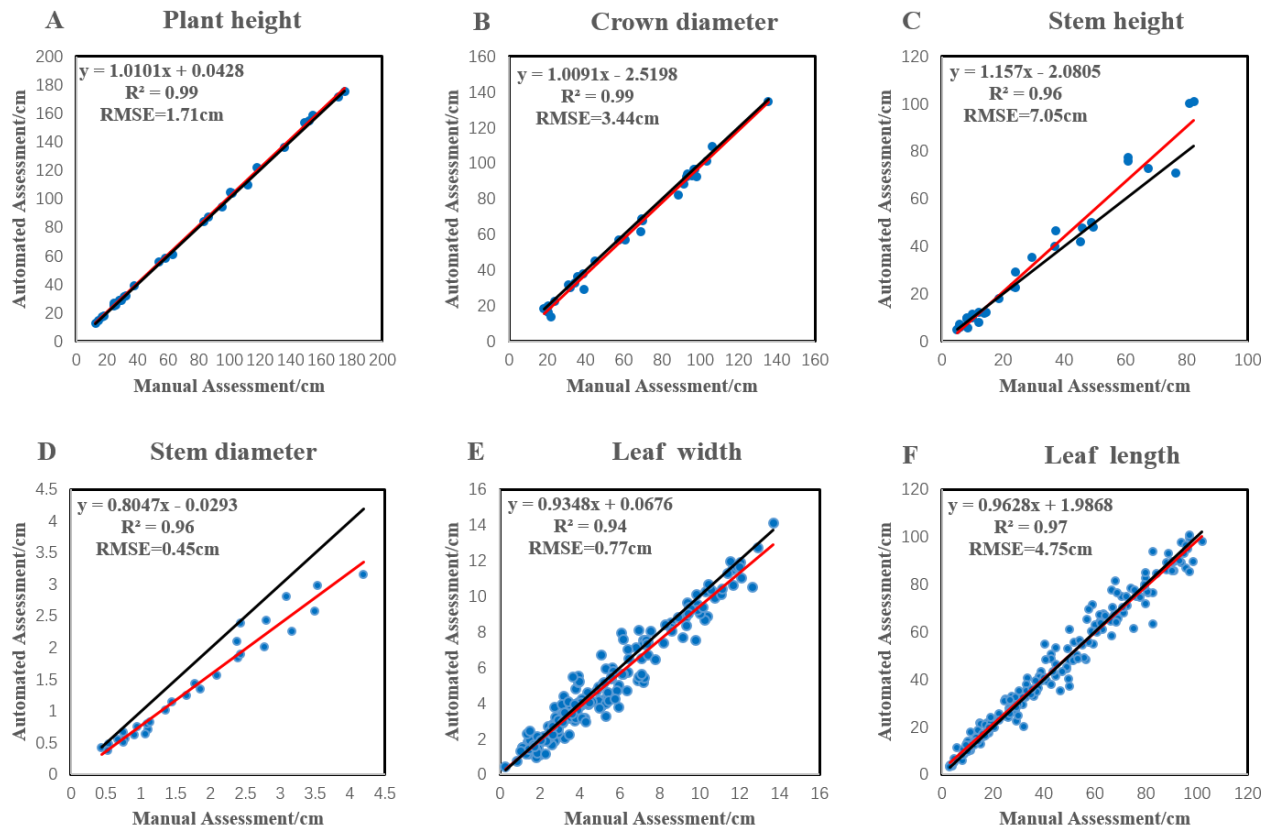

Figure 12. Comparison of six phenotyping traits extracted by our method and ground truth of the selected 30 maize samples at leaf, stem, and plant level. (A)–(F)Results of plant height, crown diameter, stem diameter, stem height, leaf width, and leaf length, respectively. Red lines indicate the least squares linear regression curves, and black lines were the idealized regression curves (y = x).

## 2.6    Results for skeleton optimization

We compared the plant skeleton generated in this paper with that generated by Laplacian based contraction method. Figure 13 shows the comparison results of the two algorithms. Our skeleton algorithm was the optimization of Laplacian based contraction method. For maize plants with different leaf numbers, our algorithm can extract the skeletons of the upper leaves completely and clearly, which is not achieved by Laplacian based contraction method. In the supplementary, we showed the skeleton visualization results of all 30 samples in detail.

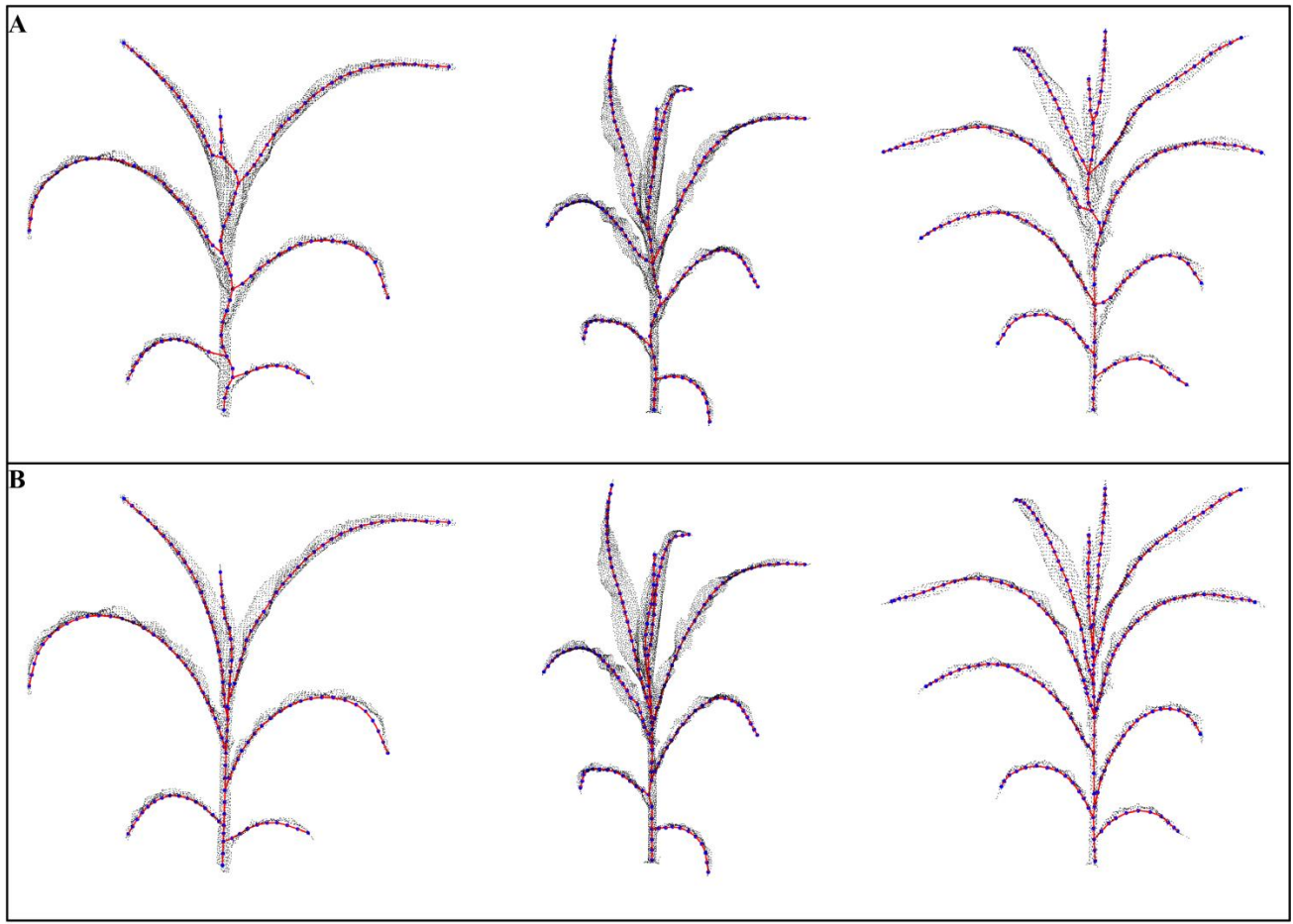

Figure 13. Comparison of the skeleton extracted by Laplacian based method and the skeleton optimized by our method. (A) The skeleton generated by Laplacian based method. (B) The optimized skeleton.

## 3    Discussion

## 3.1    Stem-leaf segmentation

### 3.1.1 Parameters

Our algorithm performed well for different maize. The reason is that the parameters of our algorithm are adjustable. The parameters are used to optimize the segmentation results in the three key stages of skeleton extraction, stem point cloud extraction and stem-leaf classification, which makes our method very flexible. The selection of these parameters is not completely random, we found that for the test 30 samples, these parameters were relatively stable and had certain rules, so users can easily find the appropriate parameters.

**Parameters for skeleton extraction**

We found by experiment that the default parameters used by Cao et al. (2010) can provide a good skeleton for most plants, except for the parameter $k$. $k$ can affect the skeleton extraction of new emerging leaves. As the value of k increases, more local features of the point cloud are ignored in the skeleton, the skeleton of the new emerging leaflet may not be extracted (Figure 13). However, if $k$ is too small, a lot of triangles will be formed in the initial skeleton, and these triangles need to be deleted by edge collapse operation, which greatly increased the skeleton extraction time. Therefore, there seems to be some contradiction in the selection of $k$ value. In our current implementation, if the new leaflet of a plant is close to other leaves, we set k to [6, 10] to facilitate complete skeleton extraction. For plants without the new leaflets, we set k to 16. In this paper, the number of point clouds of different sizes of plants was similar; therefore, the strategy of adjusting parameter k is effective for different sizes of maize plant sample.

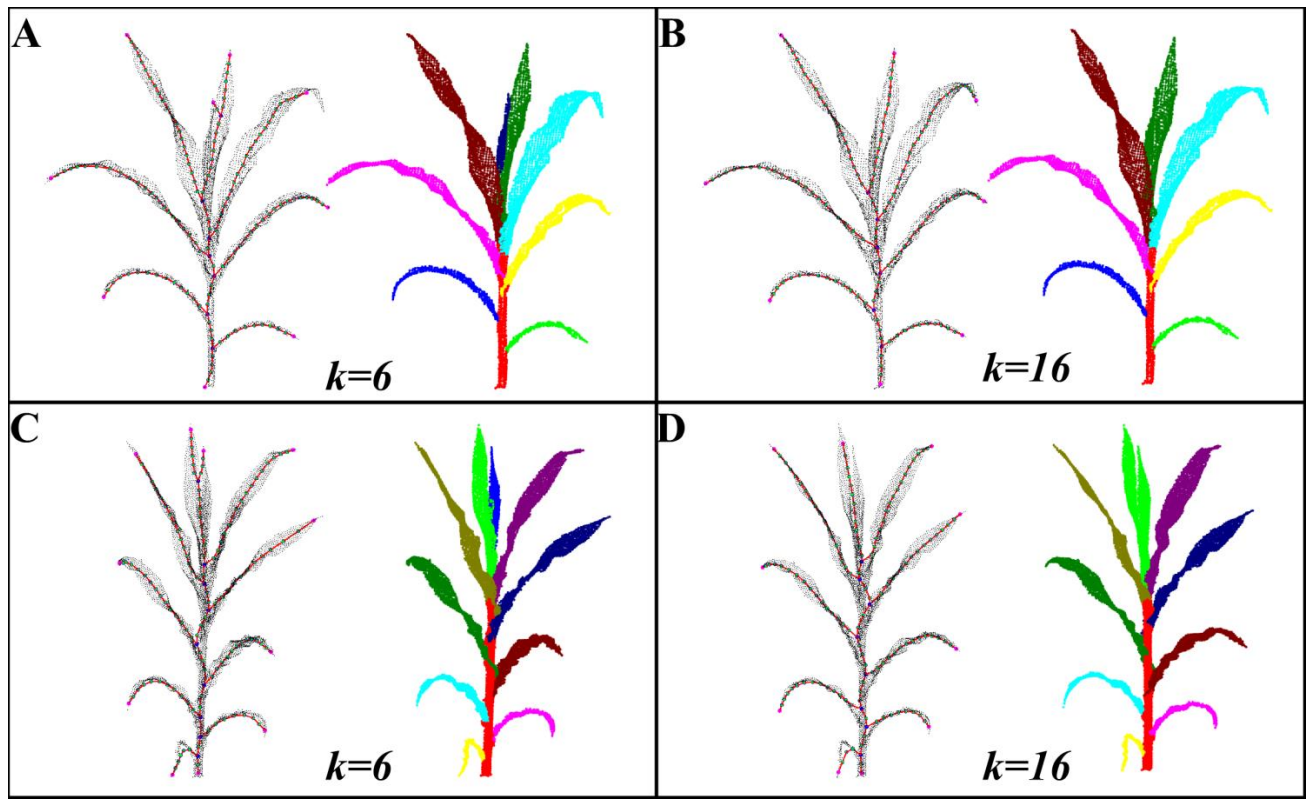

Figure 13. Influence of $k$ value on skeleton extraction and point cloud segmentation. (A),(B) represent the results of the same sample under k=6 and k=16. (C),(D) represent the results of the same sample.

## Parameters for finding stem points

In section 2.3.2, we used radius constraint to finding stem points from the set $\emptyset_u$. Parameter $\alpha$ has influence on the result of stem segmentation. The larger parameter $\alpha$ is set, the longer and thicker the stem instance will be. On the contrary, it will be shorter and thinner. For most plants at seedling stage, $\alpha = 1.0$ can obtain good stem segmentation effect (Figure 13(A)). When the plants are larger, their fully expanded leaves have obvious leaf collars on the stem. For such plants, we suggest that $\alpha$ value be set smaller (less than 1.0), so that part of the leaf collar can be segmented (Figure 13(B)). But too small $\alpha$ value will lead to under segmentation of the stem instance.

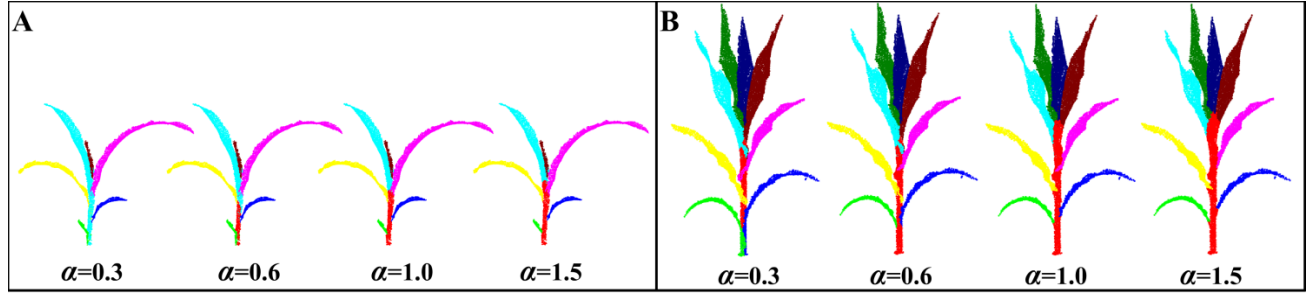

Figure 14. Segmentation results with different $\alpha$ values. (A) represents a smaller sample at seedling stage. (B) represents a larger sample with obvious leaf collars on the stem.

## Parameters for stem-leaf classification

There are two key parameters ($K_1$ and $\beta$) in our stem-leaf classification algorithm. In the fine segmentation, a point is classified according to the similarity of geometric features (Euclidean distance and plane distance) between this point and the point cloud in the local area of the organ. The parameter $K_1$ affects the size of the local area of an organ involved in the classification. When the point density is the same, $K_1$ is directly proportional to the size of the local area. In this paper, $K_1$ is set to 5 by default. The parameter $\beta$ determines the proportion of the plane distance constraint ($C_2$ term in formula (2)) in the organ distance ($C$ term in formula (2)). In this paper, $\beta$ is set to 1.0 by default.

### 3.1.2 Advantages

The proposed stem-leaf segmentation approach is fully automatic, and shows that skeleton information as prior knowledge can assist in the automation of point cloud segmentation. The prior knowledge of stem-leaf segmentation task can be obtained in many ways. Machine learning method (McCormick et al., 2016; Sodhi et al., 2017) uses a large number of manually labeled data to construct classification or segmentation models to segment plant organs. These models can be regarded as a kind of prior knowledge. Jin et al. (2018) used the manually labelled stem point cloud as a prior knowledge to realize automatic segmentation. These studies need manual data annotation when acquiring prior knowledge. In contrast, skeleton is an effective way to automatically acquire prior knowledge without manual work. Several researchers sliced the point cloud along the growth direction to generate the plant skeleton, and then segmented the organs automatically. The premise of such methods was that the point cloud was aligned along the growth direction. Xiang et al. (2019) used Kinect camera to obtain the plant point cloud, ensuring that the growth direction of the plant coincided with the z axis. Using the point cloud data obtained from multi-view images, color information data can be used to extract markers to assist in estimating the direction of plant growth, such as ground (Zermas et al., 2020) or pot (Wu et al., 2019). These studies have proved that the point cloud obtained by Kinect camera or image-based method can be used to generate the plant skeleton by slice-based method to assist the segmentation of plant organ instances. However, some point cloud data obtained by hand-held 3D scanners do not align the growth direction with the coordinate axis, nor do they have color information. Therefore, it is difficult to extract the skeleton using slicing-based methods. Our method can directly extract the skeleton of the point cloud without alignment and color information, and use only the topological and morphological features of the skeleton to recognize the organ category and estimate growth direction. This enables automation of the entire process. Therefore, our method can be used as a general automatic skeleton based stem-leaf segmentation method of maize plant without considering whether the point cloud data is registered or contains additional color information.

Our method can segment new emerging leaves that are close and wrapped together. The plant skeleton and stem-leaf classification methods play a crucial role in enabling this segmentation. The skeleton helps to distinguish the tip regions of new leaves, which is the premise of our segmentation. However, the skeleton alone is not enough to segment new leaves. Wu et al. (2019) also used the Laplacian method for maize skeleton extraction but could not obtain the complete skeleton of the new leaf, because they did not consider the geometric feature of point cloud. We employed a simple but effective stem-leaf classification method based on the tip point clouds of new leaves, obtained from the skeleton. Our stem-leaf classification method uses top-down order to classify points based on spatial consistency and geometric consistency, which is equivalent to adopting a growth region segmentation strategy from the leaf tip to the stem. The top-down order is critical as it facilitates the complete utilization of the segmented point cloud information in the new leaf while classifying the points. At the same time, it also ensures that the points belonging to the new leaves are determined first. Jin et al. (2018) used the regional growth method to segment the stem first and then segmented the leaves from the intersection of stem and leaf to leaf tip. This strategy is not suitable for new emerging leaves. The new leaves and stem are usually very close to each other, and the leaf base of the larger leaf covers that of the smaller leaf. This makes it difficult to discern which new leaf a point belongs to by analyzing the segmented stem point cloud. We used the segmented point cloud of the leaf to ascertain whether an unsegmented point belonged to the leaf, making full use of its geometric features.

Clustering algorithms are often used in point cloud segmentation of maize plants. Elnashef et al. (2019) used the DBSCAN clustering algorithm to segment individual leaves. Slice-based skeleton methods (Xiang et al., 2019; Zermas et al., 2020) use the Euclidean clustering technique (Rusu and Cousins, 2011) to generate skeleton vertices in each slice. In essence, these clustering methods use Euclidean distance for point classification. When the distance between organ point clouds is large, these methods are very robust, but it is difficult for them to find a suitable set of parameters to separate new leaves correctly, because these leaves are close together and wrapped, as described above. We compared the results of our stem-leaf classification method, Euclidean clustering technique and DBSCAN clustering algorithm for new leaf segmentation (as shown in Figure 15). We used a plant point cloud with the fully expanded leaves removed, including only the stem and four upper new leaves wrapped around each other. Compared with ground truth（Figure 15(D)）, our method perfectly segment the upper new leaves(Figure 15(A)).Figure 15(B) and Figure 15(C) are the results using Euclidean clustering technique and DBSCAN clustering algorithm respectively. We used two methods to cluster only the four new leaves, and the stem point cloud in the two subplot is the ground truth. In the two subplot, the right figure is the result for the original point cloud, the left figure is the result for the original point cloud of all the points of leaf bases are removed. The leaf base parts of new leaves are very close to each other, and the leaf base of the larger leaf covers that of the smaller leaf. Neither two clustering methods can segment such leaves correctly. After deleting the leaf base, the distance between the remaining leaf point clouds becomes large, and the two methods can separate such leaves. Compared with the two clustering methods, our stem-leaf classification is better in the segmentation of new leaves

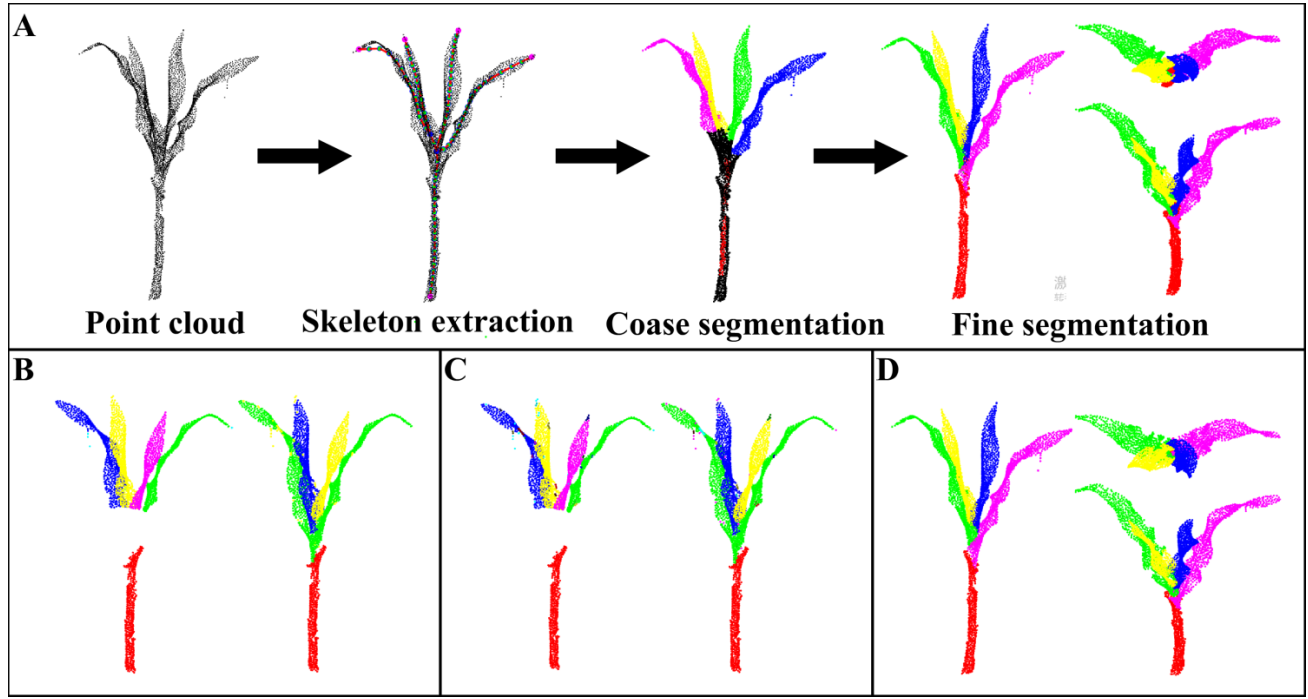

Figure 15. Comparison of new leaves segmentation results of our method, Euclidean clustering method and DBSCAN clustering method. (A) represents the process and results of segmentation using our method. (B) represents the results of segmentation using Euclidean clustering method. (C) represents the results of segmentation using DBSCAN. (D) represents the ground truth of the test sample.

We also compare our method with the regional growth method in PCL library (Rusu and Cousins, 2011). For each sample, we used exhaustive strategy to adjust all parameter values in the regional growth method, and selected the result with the largest overall accuracy value to compare with our method. Table 3 shows the accuracy comparison between our method and the regional growth method. Due to the space limitation, the detailed segmentation results of all sample using regional growth method are attached in the supplementary, including the visualization results, accuracy, recall, micro f1 score, overall accuracy.

We proved the effectiveness of our approach through experiments. While obtaining the ground truth data manually, we found it difficult to segment new leaves. Compared with manual segmentation, our method is much faster in obtaining similar segmentation results. It can therefore be used to make high-precision organ datasets for evaluating segmentation methods. Because we segment leaves completely, we can dynamically monitor the entire growth process of the leaf. Also, our stem-leaf classification strategy can be easily integrated with other point cloud segmentation methods based on the skeleton, to improve the ability of these methods in new leaf segmentation.

**Table 3  Comparison of segmentation results between our method and that of regional growth method**

| | Micro F1 Score | | | Precision | | | Recall | | | Overall Accuracy | | |
|---|---|---|---|---|---|---|---|---|---|---|---|---|
| | Min | Mean | Max | Min | Mean | Max | Min | Mean | Max | Min | Mean | Max |
| Our Method | 0.926 | 0.963 | 0.989 | 0.935 | 0.964 | 0.988 | 0.911 | 0.966 | 0.992 | 0.931 | 0.969 | 0.992 |
| Regional Growing | 0.527 | 0.739 | 0.880 | 0.479 | 0.750 | 0.887 | 0.585 | 0.731 | 0.872 | 0.639 | 0.807 | 0.899 |

## 3.2  Phenotypic trait extraction

At plant level, all the traits were independent of the point cloud segmentation, the results were remarkably close to the ground truth. Further, this also indicates that our plant coordinate system is suitable to represent the morphological structure of plants.

There were two reasons that may cause errors in leaf width estimation. The first was that two endpoints of the leaf width path were not symmetrical according to the leaf vein, so the shortest path between these points was not perpendicular to the vein, resulting in a slightly larger width. The second reason was that we took the point cloud in the middle of the leaf to calculate the leaf width, but this part may not be the widest part of the leaf, resulting in a slightly smaller width.

The error of leaf length was mainly due to the fact that leaf length path may not coincide with the vein completely. There may be three reasons for this. The first was that if the tip of a leaf drooped severely, the endpoint of its first principal component direction was not in the tip area, hence the tip part was not added to the leaf length path. The second was that the collars of some leaves were not accurately segmented, so the points of collar region were not added to the leaf length path. The last reason was that some leaf instances were over segmented and the points on the stem were wrongly added to length path. The first two reasons led to the underestimate of leaf length, and the last one led to the overestimation of leaf length.

The stem height was overestimated. Sometimes the new leaf point mistakenly regarded as the stem point. Although only few such points might exist, they were involved in the calculation of stem height, resulting in overestimation. On the contrary, the stem diameter was slightly underestimated. For estimating the stem diameter, we used the median distance from all the stem points to the middle axis of the stem as the radius of the stem, which implies that the cross section of the stem is a circle. Occasionally, the cross section of the stem was closer to the ellipse, and the median value was closer to the length of the short axis of the ellipse. However, when it was measured manually, the stem diameter indicated the length of the long axis in the ellipse. Therefore, the stem diameter was underestimated.

Although there was some underestimation or overestimation of phenotypic traits, the observed high correlations with manual measurements validated the accuracy and utility of our phenotypic trait extraction process.

### 3.3    Skeleton optimization

Our skeleton optimization algorithm can extract the skeletons of the upper leaves completely and clearly, which is not achieved by Laplacian based contraction method. Phenotypic parameters can be directly analyzed by a complete skeleton. In addition to estimate phenotypic parameters for genotype-to-phenotype studies, plant skeleton also plays an important role in many virtual agricultural applications, such as geometric modeling, shape deformation, and growth simulation (Wade and Parent, 2002; Yan et al., 2008). Therefore, this method can also be used as a skeleton generation method in these applications.

### 3.4    Limitations and Future works

The proposed method has some limitations. Firstly, the accuracy of stem segmentation was not as good as that of leaves. This is caused by two factors. The first is that our approach can not be very accurate to determine the boundary between the top of the stem and the leaf (Figure 8(B)), although sometimes it is difficult for even people to determine the boundary clearly. The second is that our method also can not extract all the leaf collar point clouds completely (Figure 8(H)), and some of them will be mistakenly segmented into stem instances. In the future, we will try to get the boundary features of stem top and leaves by machine learning.

Secondly, our algorithm cannot consider the point cloud segmentation of male and female maize ears. The Laplacian based method shrinks tassels and female ears into multiple sub-skeletons, and the current sub-skeleton decomposition method interprets their skeletons as leaf sub-skeletons. In future, we hope to include point cloud features for distinguishing ears and leaves.

Finally, if an upper leaf is squeezed or bent, its tip area cannot be separated from the other leaves; hence, our method cannot segment such leaves. Our future investigations will focus on improvements in the proposed method to make it more robust and effective in such situations.

### 4    Conclusion

In this study, we presented an automatic stem-leaf segmentation method for maize plant based on skeleton. Our method can directly extract the skeleton of the point cloud without alignment and color information, and use only the topological and morphological features of the skeleton to identify the number and category of organs which ensures that the entire algorithm is fully automated. Our stem-leaf classification method uses top-down order to classify points based on spatial consistency and geometric consistency which is particularly suitable for the segmentation of new emerging leaves. It separated leaves close to each other, and segmented effectively when large leaves wrapped small ones. Our method achieved high segmentation accuracy. The mean values of precision, recall, micro F1 score and over accuracy of each individual maize were 0.964, 0.966, 0.963 and 0.969. Using the segmentation results, two applications were also developed in this paper, including phenotypic trait extraction and skeleton optimization. Six phenotypic parameters can be accurately and automatically measured, including plant height, crown diameter, stem height and diameter, leaf width and length. Furthermore, the values of $R^2$ for the six phenotypic traits were all above 0.94. We also proposed a skeleton optimization method which can extract the skeletons of the upper leaves completely and clearly. Accordingly, we believe that the proposed algorithm can serve as a useful reference to develop automatic tools for maize phenotypic analysis and virtual agriculture application.

## 5    Conflict of Interest Statement

The authors declare no conflict of interest relating to this work.

## 6    Author Contributions

C.Z. performed the experiments, acquired the 3D data of maize and wrote the paper; T.M. wrote the programs and drafted the initial manuscript; T.Y. and T.X. conceived and designed the experiments; N.L. improved the approach in some details.

## 7    Funding

This work was supported by the China Postdoctoral Science Foundation [grant number 2018M631821], Science and Technology Program of Liaoning, China [grant number 2019JH2/10200002], and National Natural Science Foundation of China [grant number 31501217].